\documentclass[preprint,10pt]{elsarticle} 
\usepackage{graphicx}
\usepackage{url}
\usepackage{dcolumn}
\usepackage{mathtools}
\usepackage{algorithm}
\usepackage{algpseudocode}
\usepackage{amsmath,amssymb,amsfonts}

\usepackage{multirow}%
\usepackage{amsthm}%
\usepackage{mathrsfs}%
\usepackage{xcolor}%
\usepackage{textcomp}%
\usepackage{manyfoot}%
\usepackage{booktabs}%
\usepackage{listings}%
\def\vector#1{\mbox{\boldmath $#1$}}

\begin{document}

\title{ Application of linear regression and quasi-Newton methods to the deep reinforcement learning in continuous action cases }

\author{Hisato Komatsu 
}
\ead{hisato-komatsu@biwako.shiga-u.ac.jp}

\affiliation{organization={Data Science and AI Innovation Research Promotion Center, Shiga University},
postcode={522-8522},
city={Shiga},
country={Japan}}



\begin{abstract}
The linear regression (LR) method offers the advantage that optimal parameters can be calculated relatively easily, although its representation capability is limited than that of the deep learning technique. To improve deep reinforcement learning, the Least Squares Deep Q Network (LS-DQN) method was proposed by Levine \textit{et al.}, which combines Deep Q Network (DQN) with LR method. However, the LS-DQN method assumes that the actions are discrete.
In this study, we propose the Double Least Squares Deep Deterministic Policy Gradient (DLS-DDPG) method to address this limitation. This method combines the LR method with the Deep Deterministic Policy Gradient (DDPG) technique, one of the representative deep reinforcement learning algorithms for continuous action cases. For the LR update of the critic network, DLS-DDPG uses an algorithm similar to the Fitted Q iteration, the method which LS-DQN adopted. In addition, we calculated the optimal action using the quasi-Newton method and used it as both the agent's action and the training data for the LR update of the actor network.
Numerical experiments conducted in MuJoCo environments showed that the proposed method improved performance at least in some tasks, although there are difficulties such as the inability to make the regularization terms small.
\end{abstract}

\begin{keyword}
reinforcement learning \sep linear regression \sep continuous action
\end{keyword}


\maketitle

\section{Introduction \label{introduction} }
Recently, studies on deep reinforcement learning (DRL) have advanced rapidly, similar to other machine learning methods using deep neural networks. Since the introduction of the Deep Q Network (DQN), the first successful example of DRL algorithm \cite{Mnih15}, DRL methods have outperformed conventional reinforcement learning (RL) methods \cite{Mnih16, Schulman17,Lillicrap15,Fujimoto18,Haarnoja18_1,Haarnoja18_2}. However, training deep learning models requires high computational costs, especially in RL, where agents must gather data by themselves. Therefore, improving the efficiency of DRL methods has become crucial.

The linear regression (LR), one of the earliest machine learning methods, has a lower representation capability compared to deep learning. However, it offers the advantage of calculating optimal parameters with relatively low computational cost. In the case of neural networks (NNs), if the activation function of the output layer is linear, the output weight matrix can be trained by LR, using the last hidden layer as explanatory variables. 

To improve the sampling efficiency and performance of DRL, Levine \textit{et al.} proposed the Least Squares Deep Q Network (LS-DQN) method \cite{Levine17}. In this method, two types of updates are performed: updating the whole NN using the DQN and updating the output weight matrix using LR. Their study demonstrated that the LS-DQN method recorded higher scores than the original DQN method in certain Atari games. However, LS-DQN assumes that the action has discrete values like the original DQN, and therefore cannot be applied to environments with continuous actions. Another approach to utilize LR method in DRL is the Two-Timescale Network (TTN) \cite{Chung19}. This algorithm combines LR method with representation learning to evaluate the value function, $V(s)$. The idea of using the representation learning is also reflected in the Least Squares Deep Policy Gradient (LS-DPG) method \cite{Li24}. LS-DPG is an on-policy algorithm that combines the actor-critic method, representation learning, and the LR method. The LS-DPG method is theoretically applicable to continuous action cases in principle, although numerical experiments on such cases have not conducted yet. However, considering that on-policy and off-policy algorithms have different advantages, it would be valuable to develop an off-policy algorithm like LS-DQN for continuous action scenarios.

In this study, we propose the Double Least Squares Deep Deterministic Policy Gradient (DLS-DDPG) method, which combines LR with the Deep Deterministic Policy Gradient (DDPG) method, a representative DRL algorithm for continuous action scenarios \cite{Lillicrap15}. 
The DDPG method uses two NNs: actor and critic. The actor determines the appropriate action based on the current state, while the critic evaluates the action-value function $Q(s,a)$, similar to the NN in the DQN method. DLS-DDPG method trains both of these two NNs by combining the DDPG and LR methods. For the update of the critic, we used calculations similar to that of LS-DQN. Additionally, the output of the actor at each time step is modified using the quasi-Newton method, and the calculated optimal action is also employed as the training data for the LR update of the actor.

The remainder of this paper is organized as follows: Sec.~\ref{related_work} reviews related previous studies, Sec.~\ref{method} presents our proposed method, Sec.~\ref{experiments} discusses the results of numerical simulations, and Sec.~\ref{summary} summarizes the study.

\section{Related work \label{related_work} }

\subsection{Deep Deterministic Policy Gradient (DDPG) method \label{review_DDPG} }

Some RL algorithms, such as traditional Q-learning \cite{Watkins,Sutton} and DQN, determine the policy using a greedy method, with the exception of the exploration noise. This method selects the action that maximizes the action-value function $Q(s, a)$ as the optimal action. However, in the cases where the action space is continuous, finding such maximum is difficult, because the number of possible actions is infinite. Therefore, these algorithms are rarely utilized in continuous action cases.

DDPG method is one of the representative DRL algorithms for continuous action cases. It is a variant of the Deterministic Policy Gradient (DPG) method \cite{Silver14} that incorporates deep learning. This method uses two NNs: the actor, which determines the agent's policy, $\mu (s)$, and the critic, which evaluates the action-value function, $Q(s, a)$. One key feature of DDPG is that it trains parameters using an off-policy method. This allows the utilization of a replay buffer, which enhances sampling efficiency. Additionally, unlike some other representative actor-critic algorithms, such as the Advantage Actor Critic (A2C) \cite{Mnih16} and Proximal Policy Optimization (PPO) \cite{Schulman17}, the critic in DDPG evaluates the action-value function $Q(s,a)$, rather than the value $V(s)$.
Specifically, the loss function for the critic in DDPG is given as follows: 
\begin{equation}
L_c (\varphi) = \frac{1}{N_{ \mathrm{mb} } } \sum _{t : \mathrm{minibatch} } \left( Q_{\varphi} \left( s_t , a_t \right) - y_t \right) ^2 , \label{critic_DDPG1}
\end{equation}
\begin{equation}
\mathrm{where} \ \ y_t \equiv r_t + \gamma (1-d_t) Q_{\varphi ^{targ}} \left( s' _t , \mu _{\theta ^{targ} } (s' _t) \right) . \label{critic_target}
\end{equation}
Here, $N_{ \mathrm{mb} }$ represents the minibatch size, and $\theta ^{targ}$ and $\varphi ^{targ}$ are parameters of the target networks of the actor and critic, respectively. This loss function is similar to that of DQN. The actor in DDPG is updated using the following gradient ascent:
\begin{equation}
\nabla _{\theta} \frac{1}{N_{ \mathrm{mb} } } \sum _{t : \mathrm{minibatch} } Q_{\varphi} \left( s _t , \mu _{\theta} (s_t) \right) , \label{actor_DDPG_grad}
\end{equation}
This operation is equivalent to treating the loss function of the actor as follows: 
\begin{equation}
L_a (\theta) = \frac{1}{N_{ \mathrm{mb} } } \sum _{t : \mathrm{minibatch} } -Q_{\varphi} \left( s _t , \mu _{\theta} (s_t) \right) . \label{actor_DDPG1}
\end{equation}
DDPG typically adopts the soft target updates, which gradually update the target networks after each update of the main networks using following equations:
\begin{eqnarray}
 \varphi ^{targ} & \leftarrow & \left( 1 - \tau _{\mathrm{DDPG} } \right) \varphi ^{targ} + \tau _{\mathrm{DDPG} } \varphi , \label{DDPG_softc} \\
 \theta ^{targ} & \leftarrow & \left( 1 - \tau _{\mathrm{DDPG} } \right) \theta ^{targ} + \tau _{\mathrm{DDPG} } \theta . \label{DDPG_softa}
\end{eqnarray}
This technique stabilizes the learning.

DDPG is one of the fundamental off-policy DRL algorithms for continuous action cases, and several improved variants, such as the Twin Delayed DDPG (TD3) \cite{Fujimoto18} and Soft Actor Critic (SAC) \cite{Haarnoja18_1,Haarnoja18_2}, have been proposed. To explore the potential combination of LR with these improved algorithms in the future, it is important to study whether the LR update can improve DDPG.

\subsection{Linear regression (LR) applied to reinforcement learning (RL) \label{review_LR} }

RL algorithms using LR were studied before the origin of DRL algorithms. While the representation capabilities of these algorithms are inferior to those of deep learning, they remain useful in some situations because they can calculate the optimal parameters for batches relatively easily.
The simplest way to apply linear approximation to RL is to approximate the action-value function $Q(s,a)$ as a linear combination of given functions of $s$ and $a$, $\vector{\phi} (s,a) = \left( \phi _1 (s,a) , ... , \phi _N (s,a) \right) $:
\begin{equation}
Q(s,a) = \sum _{i=1} ^N w _i \phi _i (s,a) = \vector{w} \vector{\phi} (s,a) ^T , \label{Q_LR}
\end{equation}

Note that $\vector{\phi}$ and $\vector{w}$ are row vectors. Unlike in supervised learning, where the parameters $\vector{w}$ can be easily trained using previously given data, RL training must be executed iteratively with exploration. Therefore, several algorithms have been proposed to train $\vector{w}$ in eq.~(\ref{Q_LR}).
One such method is the Least Squares Temporal Difference-Q (LSTD-Q) method proposed by Lagoudakis and Parr \cite{LP03}, which calculate $\vector{w}$ as:
\begin{equation}
\vector{w} = b_{\mathrm{LSPI} } A_{\mathrm{LSPI} } ^{-1} , \label{w_LSPI}
\end{equation}
where 
\begin{eqnarray}
A_{\mathrm{LSPI} } & = & \sum _{t : \mathrm{batch} } \left( \vector{\phi} (s_t , a_t ) - \gamma \vector{\phi} (s_{t+1} , \pi (s_{t+1} ) ) \right) ^T 
\vector{\phi} (s_t , a_t ) , \\
b_{\mathrm{LSPI} } & = & \sum _{t : \mathrm{batch} } r_t \vector{\phi} (s_t , a_t ) .
\end{eqnarray}
Here, $s_t, a_t,$ and $r_t$ represent the state, action and reward at time step $t$, and $\gamma \in (0,1)$ is the discount factor.  
The method that updates $\vector{w}$ and the corresponding policy iteratively using the LSTD-Q method is called as the Least Squares Policy Iteration (LSPI) method.

Ernst \textit{et al.} proposed another RL algorithm called Fitted Q Iteration (FQI) \cite{Ernst05}. This method minimizes the following loss function using a regression algorithm:
\begin{equation}
L_{\mathrm{FQI} } = \frac{1}{\sum _{t : \mathrm{batch} } 1 } \sum _{t : \mathrm{batch} } \left( Q \left( s_t , a_t \right) - \tilde{y}_t \right) ^2 , \label{loss_FQI1}
\end{equation}
\begin{equation}
\mathrm{where} \ \ \tilde{y} _t \equiv r_t + \gamma \max _{a} Q_{\mathrm{temp} } ( s_{t+1} , a ) \label{targ_FQI}
\end{equation}
Here, $Q_{\mathrm{temp} }$ represents the current estimation of $Q$. In principle, arbitrary functional forms can be used for the function approximation of $Q$. Comparing eqs.~(\ref{critic_DDPG1}) and (\ref{loss_FQI1}), this algorithm is similar to DDPG and DQN in terms of the loss function, and $Q_{\mathrm{temp} }$ corresponds to the target network. Conversely, DDPG and DQN can be considered deep learning variants of FQI. If the linear approximation given by eq.~(\ref{Q_LR}) is introduced, $\vector{w}$ is calculated using the following equations:
\begin{equation}
\vector{w} = b_{\mathrm{FQI}} A _{\mathrm{FQI}} ^{-1} , \label{w_FQI}
\end{equation}
where 
\begin{eqnarray}
A_{\mathrm{FQI}} & = & \sum _{t : \mathrm{batch} } \vector{\phi} (s_t , a_t ) ^T \vector{\phi} (s_t , a_t ) , \label{A_FQI} \\
b_{\mathrm{FQI}} & = & \sum _{t : \mathrm{batch} } \tilde{y} _t \vector{\phi} (s_t , a_t ) . \label{B_FQI} 
\end{eqnarray}
In the following, the word FQI represents the update in accordance with eqs.~(\ref{w_FQI})--(\ref{B_FQI}).

As the basis for the linear approximation in the above algorithms, $\vector{\phi} (s,a)$, we can use arbitrary functions. Therefore, by letting $\vector{\phi} (s,a)$ represent the values of the neurons in the last hidden layer and coefficient vector $\vector{w}$ represent the output weight matrix, these algorithms can be applied to NNs. In this case, LR method only trains the output weight matrix. The LS-DQN method proposed by Levine \textit{et al.} alternates between training the entire NN using DQN and updating the output weight matrix using the LR update. This algorithm improved the scores of some Atari games compared to DQN, no matter whether LSPI or FQI was employed as the LR update. As the NNs that train the output weight matrix using LR, Extreme Learning Machine (ELM) \cite{Huang04} and Echo State Network (ESN) \cite{Jaeger01} also exist. These networks have the same architectures as multi-layer perceptron and recurrent NN, respectively, but the weight matrices, except for the output one, are fixed with random values. The LR methods explained above have also been applied to such specialized NNs \cite{LZXZRFL19,ZLSZ21,Komatsu23}.

In these algorithms, the optimal action at each state $s$ is evaluated as $\mathrm{argmax} _a Q (s,a)$, as in Q-learning and DQN. However, calculating this argmax becomes difficult when the action space is continuous, as there are an infinite number of possible actions. Most of the recent DRL methods for continuous action cases use actor-critic algorithms, which can train an appropriate policy without requiring such a calculation. However, the policy gradient theorem, which is used to calculate the gradient ascent of the actor, cannot generate the objective function suitable for LR method. Indeed, in the case of DDPG, for example, LR method cannot be applied to eq.~(\ref{actor_DDPG1}), even if the trainable parameters are restricted to the output weight of the actor. Hence, to combine LR with actor-critic methods, most of previous studies have applied LR only to the critic, regardless of whether the DRL methods were introduced \cite{Li24,OKKhP11}. 

\section{Proposed method \label{method} }

In this study, we propose the DLS-DDPG method, which is based on DDPG and utilizes the LR update for both actor and critic. DDPG is similar to DQN in that it evaluates action-value function $Q(s,a)$ using an off-policy method. Therefore, it is expected that the concepts from LS-DQN can be directly extended, compared to other representative actor-critic algorithms. 

In the original DDPG, the output of the actor is used directly as the agent's action after adding exploration noise. However, in our approach, we calculated the optimal action, $o$, using the quasi-Newton method. Specifically, we adopted the L-BFGS-B algorithm \cite{BLNZ95} and used \texttt{scipy.optimize.fmin\_l\_bfgs\_b} for the actual implementation. Note that we assigned $(-Q)$ to \texttt{fmin\_l\_bfgs\_b} because this method minimizes, rather than maximizes, the assigned function. As the hyperparameters for the L-BFGS-B algorithm other than the maximum number of iterations, default values provided by the library were used.
As explained in Sec.~\ref{review_DDPG}, finding the argument that maximizes a continuous function is difficult. The quasi-Newton method is a frequently used algorithm to solve this problem, but its result depends on the initial value. Therefore, we should use a good approximation of the true optimal action as the initial value. In this study, for this initial value, we used the clipped output of the actor:
\begin{equation}
\mu _{\theta} (s) = C( \mu _{0,\theta} (s) ) \equiv clip( \mu _{0,\theta} (s), a_{\mathrm{low}}, a_{\mathrm{high}} ) , \label{action_clip1}
\end{equation}
where $ a_{\mathrm{high}} $ and $ a_{\mathrm{low}}$ represent the upper and lower bounds of the action allowed by the environment, and $\mu _{0,\theta}$ is the output of the actor. 
The upper and lower bounds for the L-BFGS-B algorithm, $u _{\mathrm{qN} }$ and $l _{\mathrm{qN} }$, were set as follows:
\begin{equation}
u _{\mathrm{qN}} = C( \mu _{\theta} (s) + b ) , \ \ l _{\mathrm{qN}} = C( \mu _{\theta} (s) - b  ) .
\label{bound_lbfgsb}
\end{equation}
By introducing these bounds, the difference between each component of $o$ and $\mu _{\theta} (s)$ was kept below the hyperparameter $b$. This ensures that the calculated optimal action does not differ significantly from the prediction of the actor.
In the following, we denote the value $o$ calculated by the above procedure as
\begin{equation}
o = \underset{a \in [ l _{\mathrm{qN}} ,  u _{\mathrm{qN}} ] }{ \widetilde{ \mathrm{argmax} } } \left( Q(s,a) ; \mu _{\theta} (s) \right) .
\label{optim_lbfgs}
\end{equation}
Here, $\widetilde{ \mathrm{argmax} } _a ( f(a) ; a_0) $ represents the estimated value of the argmax of $f(a)$, calculated by the L-BFGS-B method starting from $a_0$, and it may not always coincide with the true argmax.
Note that we set the output layer of the actor be linear to facilitate the LR calculation, whereas the original DDPG paper adopted a tanh layer for it to bound the range of action. Instead, the action is bounded by the clipping defined in eqs. (\ref{action_clip1}) and (\ref{bound_lbfgsb}).
For the agent's action, we added the exploration noise, $\epsilon$, to the optimal action and clipped the result:
\begin{equation}
a = C (o + \epsilon ) . \label{OAC}
\end{equation}
Additionally, the optimal action $o_t$ and the corresponding observed state, $s_t$, at time $t$ are stored in the replay buffer for LR calculation, $\mathcal{D} _{\mathrm{LR} }$. We refer to the action selection using eq.~(\ref{OAC}) as the optimal action choosing (OAC).

The update of the NNs using LR was executed every $T_{\mathrm{LR} }$ steps, and $\mathcal{D} _{\mathrm{LR} }$ was cleared after each update. 
In each update, we first constructed the matrices $O$ and $X _{a }$, where the rows were the optimal action and the last hidden layer of the actor at each time, respectively:
\begin{equation}
O = \left( 
\begin{array}{c}
o_{t = t_0}  \\
 \vdots \\
o_{t = t_0 + T _{\mathrm{LR} } -1 }
\end{array}
\right) , \ \ X _{a } = \left( 
\begin{array}{c}
x_a \left( s_{t = t_0} \right) \\
 \vdots \\
x_a \left( s_{t = t_0 + T _{\mathrm{LR} } -1 } \right)
\end{array}
\right) .
\end{equation}
Here, the value of the last hidden layer, $x_a \left( s_t \right)$, is calculated from the state $s_t$ stored in $\mathcal{D} _{\mathrm{LR} }$, using the current parameters of the NN. We did not set $T_{\mathrm{LR} }$ too large so that the stored data on the optimal action would not become outdated and unusable for the training. To stabilize the LR update using such small training data, we also used the replay buffer for DDPG, $\mathcal{D}$. Specifically, we first sampled a minibatch from $\mathcal{D}$ and calculated the last hidden layer, $X _{a,\mathrm{LRmb} }$:
\begin{equation}
X _{a,\mathrm{LRmb} } = \left( 
\begin{array}{c}
 \vdots \\
x_a \left( s_{t} \right) \\
 \vdots
\end{array}
\right) _{t \in \mathrm{LR-minibatch} } .
\end{equation}
The size of this minibatch was larger than both $T_{\mathrm{LR} }$ and the minibatch for DDPG. Considering that the actor before LR update determines the optimal action for $s_t$ as:
\begin{equation}
\theta ^{\mathrm{out}} _{\mathrm{temp} } x _a \left( s_{t} \right) ^T ,
\end{equation}
we restricted the rapid update using both $O$ and this quantity as the training data. In short, the update of the actor using LR was executed using the following equation:
\begin{equation}
\theta ^{\mathrm{out}} = b_{a} A _{a} ^{-1} , \label{DLSDDPG_actor_LR}
\end{equation}
where
\begin{eqnarray}
 A _{a} & = & X _{a} ^T X _{a} + w_a 
 \left( X _{a,\mathrm{LRmb} } ^T X _{a,\mathrm{LRmb} } + \beta _a N_{ \mathrm{LRmb} } I \right) , \label{DLSDDPG_A} \\
 b_{a} & = & O ^T X _{a} + w_a \theta ^{\mathrm{out}} _{\mathrm{temp} } 
 \left( X _{a,\mathrm{LRmb} } ^T X _{a, \mathrm{LRmb} } + \beta _a N_{ \mathrm{LRmb} } I \right) . \label{DLSDDPG_b}
\end{eqnarray}
Here, $\theta ^{\mathrm{out}} _{\mathrm{temp} }$ represents the current value of the output weight matrix of the actor.
The hyperparameter $w_a$ was introduced to control the speed of the update, and $N_{ \mathrm{LRmb} }$ was the size of the minibatch used for the LR update.
Here, we adopted Bayesian regression \cite{OHagan} and added the term $\beta _a N_{ \mathrm{LRmb} } I$ to both $A_{a}$ and $b_{a}$, because Ridge regularization empirically caused $\theta ^{\mathrm{out}} $ to become too small.

The update of the critic was similar to that of FQI method:
\begin{equation}
\varphi ^{\mathrm{out}} = b_{c} A _{c} ^{-1} , \label{DLSDDPG_critic_LR}
\end{equation}
where
\begin{eqnarray}
 A _{c} & = & X _{c,\mathrm{LRmb} } ^T X _{c,\mathrm{LRmb} } + \beta _c N_{ \mathrm{LRmb} } I , \\
 b_{c} & = & \tilde{Y} _{\mathrm{LRmb} } ^T  X _{c, \mathrm{LRmb} } , \\
 X _{c,\mathrm{LRmb} } & = & \left( 
\begin{array}{c}
 \vdots \\
x_c \left( s_{t} , a_{t} \right) \\
 \vdots
\end{array}
\right) _{t \in \mathrm{LR-minibatch} } 
\end{eqnarray}
and $\tilde{Y} _{\mathrm{LRmb} }$ represents a column vector whose $t$-th component is expressed as:
\begin{equation}
 \left( \tilde{Y} _{\mathrm{LRmb} } \right) _{t} = r_{t} + \gamma(1-d_{t} ) Q _{\varphi ^{targ} } \left( s' _{t} , \mu _{\theta ^{targ} } ( s' _{t} ) \right) \label{DLSDDPG_critic_target}
\end{equation}
To construct $X _{c,\mathrm{LRmb} }$ and $\tilde{Y} _{\mathrm{LRmb} }$, we used the same minibatch as for $X _{a,\mathrm{LRmb} }$.
Note that we treated the term $\beta _c N_{ \mathrm{LRmb} } I$ as the Ridge term and did not use Bayesian regression unlike the actor. This is because the critic is more vulnerable to the divergence of the weight matrices than the actor, and we aimed to suppress this by reducing the norm of these matrices. We did not optimize the action using the quasi-Newton method in the calculation of $Q _{\varphi ^{targ} }$ in eq.~(\ref{DLSDDPG_critic_target}). The difference between our update for the critic and the original FQI is that ours uses the target network, as shown in eq.~(\ref{DLSDDPG_critic_target}), whereas FQI uses $Q_{\mathrm{temp} }$, the current estimation of $Q$, instead. While there are such differences, FQI is more straightforward to integrate with the target network than LSPI.
After the LR updates of the NNs, we executed a soft update of the target networks:
\begin{eqnarray}
 \varphi ^{targ} & \leftarrow & \left( 1 - \tau _{\mathrm{LR} } \right) \varphi ^{targ} + \tau _{\mathrm{LR} } \varphi , \label{DLSDDPG_softc} \\
 \theta ^{targ} & \leftarrow & \left( 1 - \tau _{\mathrm{LR} } \right) \theta ^{targ} + \tau _{\mathrm{LR} } \theta . \label{DLSDDPG_softa}
\end{eqnarray}
These equations have the same form as eqs.~(\ref{DDPG_softc}) and (\ref{DDPG_softa}), but the update rate $\tau _{\mathrm{LR} }$ is different.

In summary, with regard to the critic, our algorithm uses a slightly modified FQI method for the LR update. This calculation is a straightforward extension of LS-DQN. In contrast, the output of the actor is used as the initial value of the quasi-Newton method to calculate the optimal action. This optimal action is employed both for agent's action and the training data for the LR update of the actor. Note that the quasi-Newton method is only used to optimize the action, not to update NN parameters.

We also introduced modification terms to the loss functions of DDPG. First, $L^2$-regularization terms were added to both actor and critic to prevent the divergence of the weight matrices. Additionally, we introduced a penalty term
\begin{eqnarray}
& & \frac{c}{N_{ \mathrm{mb} } D_a} \sum _{t : \mathrm{minibatch} } \left\| \mu _{0,\theta} (s_t) - \mu _{\theta} (s_t) \right\| ^2 \nonumber \\
& = & \frac{c}{N_{ \mathrm{mb} } D_a} \sum _{t : \mathrm{minibatch} } \left\| \mu _{0,\theta} (s_t) - C \left( \mu _{0,\theta} (s_t) \right) \right\| ^2 ,
\end{eqnarray}
to $L_a$, to let the actor output remains within the range permitted by the environment. Note that in this paper, the norm of a vector or matrix refers to the $L^2$- or Frobenius norm, i.e., the square root of the sum of the squares of all components. Here, $D_a$ is the dimension of the action space, and the coefficient $c$ is a hyperparameter. In summary, eqs.~(\ref{critic_DDPG1}) and (\ref{actor_DDPG1}) are modified as follows:
\begin{equation}
L_c (\varphi) = \frac{1}{N_{ \mathrm{mb} } } \sum _{t : \mathrm{minibatch} } \left( Q_{\varphi} \left( s_t , a_t \right) - y_t \right) ^2 + \beta ' _c \left\| \varphi \right\| ^2 , \label{DLSDDPG_critic_loss}
\end{equation}
\begin{eqnarray}
L_a (\theta) & = & \frac{1}{N_{ \mathrm{mb} } } \sum _{t : \mathrm{minibatch} } \biggl[ -Q_{\varphi} \left( s _t , \mu _{\theta} (s_t) \right) \biggr. \nonumber \\
& & \left. + \frac{c}{D_a} \left\| \mu _{0,\theta} (s_t) - \mu _{\theta} (s_t) \right\| ^2 \right] + \beta ' _a \left\| \theta \right\| ^2 . \label{DLSDDPG_actor_loss}
\end{eqnarray}
In eqs.~(\ref{DLSDDPG_critic_loss}) and (\ref{DLSDDPG_actor_loss}), $\varphi$ and $\theta$ represent all parameters of each NN.  

Additionally, we update the coefficients of the regularization terms, $\beta_a$, $\beta ' _a$, $\beta_c$, and $\beta ' _c$ before the LR update. Here, we first calculate the Frobenius norms of the output weight matrices normalized by the square roots of the numbers of their components, $N_{\theta ^{out}}$ and $N_{\varphi ^{out}}$:
\begin{eqnarray}
n_{\theta} & \equiv & \frac{\left\| \theta ^{out} \right\| ^2  }{ \sqrt{ N_{\theta ^{out}} } }  = \sqrt{ \frac{1}{N_{\theta ^{out}} } \sum _{ij} \left| \theta ^{out} _{ij} \right| ^2 } , \nonumber \\
n_{\varphi} & \equiv & \frac{\left\| \varphi ^{out} \right\| ^2  }{ \sqrt{ N_{\varphi ^{out}} } } = \sqrt{ \frac{1}{N_{\varphi ^{out}} } \sum_{j} \left| \varphi ^{out} _{j} \right| ^2 } , \nonumber
\end{eqnarray}
and adjust the manipulation depending on their values. Specifically, each coefficient returned to its initial value if the corresponding normalized norm was larger than a threshold value, $C_a$ or $C_c$, and was gradually decreased to its minimum value otherwise.
For example, the update of $\beta _a$ was given as follows:
\begin{equation}
\beta _a \leftarrow \left\{ 
\begin{array}{cc} 
\beta _{a,0} & \mathrm{if} \ \ n_{\theta} > C_a \\
\max \left( \delta \beta _a, \beta _{a,min} \right) & \mathrm{otherwise}
\end{array} \right. . \label{DLSDDPG_beta_a}
\end{equation}
Here, the initial and minimum values, $\beta _{a,0}$ and $\beta _{a,min}$, and the decay rate, $\delta$, were hyperparameters. The other coefficients were also updated using similar manipulations:
\begin{eqnarray}
\beta ' _a & \leftarrow & \left\{ 
\begin{array}{cc} 
\beta ' _{a,0} & \mathrm{if} \ \ n_{\theta} > C_a \\
\max \left( \delta \beta ' _a, \beta ' _{a,min} \right) & \mathrm{otherwise}
\end{array} \right. , \label{DLSDDPG_beta_prime_a} \\
\beta _c & \leftarrow & \left\{ 
\begin{array}{cc} 
\beta _{c,0} & \mathrm{if} \ \ n_{\varphi} > C_c \\
\max \left( \delta \beta _c, \beta _{c,min} \right) & \mathrm{otherwise}
\end{array} \right. , \label{DLSDDPG_beta_c} \\
\beta ' _c & \leftarrow & \left\{ 
\begin{array}{cc} 
\beta ' _{c,0} & \mathrm{if} \ \ n_{\varphi} > C_c \\
\max \left( \delta \beta ' _c, \beta ' _{c,min} \right) & \mathrm{otherwise}
\end{array} \right. . \label{DLSDDPG_beta_prime_c}
\end{eqnarray}
The entire algorithm is summarized in Algorithm \ref{alg1}.

\begin{figure}[!t]
\begin{algorithm}[H]
    \caption{DLS-DDPG}
    \label{alg1}
\begin{algorithmic}[1]
   \State $\beta _a \gets \beta _{a0}, \beta ' _a \gets \beta ' _{a0}, \beta _c \gets \beta _{c0}, \beta ' _c \gets \beta ' _{c0}$, $\mathcal{D} , \mathcal{D}_{\mathrm{LS} } \gets \emptyset$
   \State initialize parameters of NNs, $\varphi$, $\theta$, $\varphi ^{\mathrm{targ}}$ and $\theta ^{\mathrm{targ}}$ 
   \State Reset the environment and get the state $s$
   \For {$t_{ \mathrm{global} } \gets 1, ... , T_{\mathrm{max} } $}
      \If { $t_{ \mathrm{global} } > T_{ \mathrm{rand} } $ }
         \State Calculate optimal action $o$
         \State Decide agent's action $a$ by adding noise to $o$ 
      \Else
         \State Choose agent's action $a$ randomly
      \EndIf
      \State Update the environment and observe the reward $r$, next state $s'$, and done signal $d$
      \State Store $(s , a , r , s' , d )$ in the replay buffer for DDPG, $\mathcal{D}$
      \If { $t_{ \mathrm{global} } > T_{ \mathrm{rand} } $ }
         \State Store $(s , o)$ in the replay buffer for LR, $\mathcal{D}_{\mathrm{LR} }$
         \State Update $\varphi$, $\theta$, $\varphi ^{\mathrm{targ}}$ and $\theta ^{\mathrm{targ}}$ using DDPG
      \ElsIf { $t_{ \mathrm{global} } = T_{ \mathrm{rand} } $ }
         \State Update $\varphi$, $\theta$, $\varphi ^{\mathrm{targ}}$ and $\theta ^{\mathrm{targ}}$ using DDPG $T_{ \mathrm{rand} }$ times \label{DDPG_for_RDsteps}
      \EndIf 
         
      \If { $t_{ \mathrm{global} } \equiv 0 $ (mod $T_{\mathrm{LR} }$) and $t_{ \mathrm{global} } > T_{ \mathrm{rand} } $ }
         \State Update $\beta_a$, $\beta ' _a$, $\beta_c$, and $\beta ' _c$ using eqs.~(\ref{DLSDDPG_beta_a})--(\ref{DLSDDPG_beta_prime_c})
         \State Update $\varphi ^{\mathrm{out}} $ using LR given by eq.~(\ref{DLSDDPG_critic_LR})
         \State Update $\theta ^{\mathrm{out}} $ using LR given by eq.~(\ref{DLSDDPG_actor_LR})
         \State Update $\varphi ^{targ} $ and $\theta ^{targ} $ using eqs.~(\ref{DLSDDPG_softc}) and (\ref{DLSDDPG_softa})
         \State $\mathcal{D}_{\mathrm{LR} } \gets \emptyset$
      \EndIf
      \If {the environment is terminated or truncated}
         \State Reset the environment and get the state $s$
      \Else
         \State $s \gets s'$
      \EndIf
   \EndFor
\end{algorithmic}
\end{algorithm}
\end{figure}

In this study, we used NNs with one hidden layer for simplicity. Although our architecture was shallower than standard ones, we observed few problems caused by this in numerical experiments, as described in Sec.~\ref{experiments}. The activation function for the hidden layers is tanh. Hence, $Q(s,a)$ is expressed as follows:
\begin{equation}
Q(s, a) = \varphi ^{out} \left( 
\begin{array}{c}
\tanh \left( \varphi _s ^{in} \tilde{s} + \varphi _a ^{in} a \right)  \\
1 
\end{array}
\right) ,
\end{equation}
\begin{equation}
\mathrm{where} \ \tilde{s} = \left(
\begin{array}{c}
s \\
1 
\end{array}
\right) ,
\end{equation}
and the $j$-th component of the derivative of $Q$ regarding $a$ is calculated as:
\begin{eqnarray}
\left( \nabla _a Q(s, a) \right) _j & = & \sum _{i : \mathrm{hidden} } \varphi _i ^{out} \nabla _a \tanh \left( \varphi _s ^{in} \tilde{s} + \varphi _a ^{in} a \right) _i \nonumber \\
& = & \sum _{i : \mathrm{hidden} } \varphi _i ^{out} (\varphi _{a} ^{in} )_{ij} 
 \left[ 1 - \tanh ^2 \left( \varphi _s ^{in} \tilde{s} + \varphi _a ^{in} a \right) \right] _i . \label{fprime_Q}
\end{eqnarray}
Eq.~(\ref{fprime_Q}) (times (-1)) was assigned to the variable \texttt{fprime}, representing the derivative of the function, of the method \texttt{fmin\_l\_bfgs\_b}. We avoided using ReLU as the activation function because quasi-Newton methods, including the L-BFGS-B method, assume the existence of the second derivative of the function. 

\begin{table*}
\hspace{-4.0cm}
\begin{tabular}{cccc}
 character & meaning & our calculation & usual DDPG \\ \hline\hline
 -- & optimizer for the deep learning & \multicolumn{2}{c}{Adam \cite{Kingma14} } \\ \hline
 -- & the number of neurons of the hidden layer & 1024 & (400, 300) \\ \hline
 -- & activation functions of the hidden layers & tanh & ReLU \\ \hline
 -- & activation function of the output layer of the actor & linear & tanh \\ \hline
 $T_{ \mathrm{rand} }$ & initial random steps & 25000 & 10000 \\ \hline
 $\gamma$ & discount factor & \multicolumn{2}{c}{0.99} \\ \hline
 -- & distribution function of the exploration noise & \multicolumn{2}{c}{Gaussian} \\ \hline
 -- & standard deviation of the exploration noise & \multicolumn{2}{c}{0.1} \\ \hline
 -- & learning rate of DDPG & \multicolumn{2}{c}{0.001} \\ \hline
 -- & replay buffer size & \multicolumn{2}{c}{1000000} \\ \hline
 $N_{ \mathrm{mb} }$ & minibatch size for the DDPG update & \multicolumn{2}{c}{256} \\ \hline
 $N_{ \mathrm{LRmb} }$ & minibatch size for the LR update & 10000 & -- \\ \hline
 $T_{\mathrm{LR} }$ & interval between LR updates & 1000 & -- \\ \hline
 $w_a$ & weight of old parameters in eqs.~(\ref{DLSDDPG_A}) and (\ref{DLSDDPG_b}) & 2 & -- \\ \hline
 $\tau _{\mathrm{DDPG} }$ & target update rate after the DDPG update & \multicolumn{2}{c}{0.005} \\ \hline
 $\tau _{\mathrm{LR} }$ & target update rate after the LR update & 0.1 & -- \\ \hline
 
-- & maximum number of iterations for the L-BFGS-B method & 10 & -- \\ \hline
$b$ & upper bound of the change in the L-BFGS-B method & 0.4 & -- \\ \hline

 $\beta ' _{a,0}$ & initial coefficient of the regularization term for the DDPG update of the actor & 0.01 & -- \\ \hline
 $\beta '  _{a,min}$ & minimum coefficient of the regularization term for the DDPG update of the actor & 0.001 & -- \\ \hline
 $\beta ' _{c,0}$ & initial coefficient of the regularization term for the DDPG update of the critic & 0.01 & -- \\ \hline
 $\beta ' _{c,min}$ & minimum coefficient of the regularization term for the DDPG update of the critic & 0.001 & -- \\ \hline

 $\beta _{a,0}$ & initial coefficient of the regularization term for the LR update of the actor & 0.01 & -- \\ \hline
 $\beta _{a,min}$ & minimum coefficient of the regularization term for the LR update of the actor & 0.001 & -- \\ \hline
 $\beta _{c,0}$ & initial coefficient of the regularization term for the LR update of the critic & 0.01 & -- \\ \hline
 $\beta _{c,min}$ & minimum coefficient of the regularization term for the LR update of the critic & 0.001 & -- \\ \hline

 $c$ & coefficient of the penalty term in eq.~(\ref{DLSDDPG_actor_loss}) & 0.001 & -- \\ \hline
 $\delta$ & decay rate of $\beta_a$, $\beta ' _a$, $\beta_c$, and $\beta ' _c$ & 0.95 & -- \\ \hline
 $C_a$ & threshold used in eqs.~(\ref{DLSDDPG_beta_a}) and (\ref{DLSDDPG_beta_prime_a}) & 1 & -- \\ \hline
 $C_c$ & threshold used in eqs.~(\ref{DLSDDPG_beta_c}) and (\ref{DLSDDPG_beta_prime_c}) & 10 & -- \\ \hline 
\end{tabular}
\vspace{1.0mm}
\caption{Hyperparameters of this study. Here, the NNs of the usual DDPG had two hidden layers, and the first and second ones were composed of 400 and 300 neurons, respectively. We expressed this as (400,300). In addition to the differences in hyperparameters, $\varphi _s ^{in}$ and $\varphi _a ^{in}$, the input weight matrices of the critic for the state and the action, were treated as the two separate tensors in our architecture, whereas usual DDPG treated them as one tensor. }
\label{hyperparameters}
\end{table*}
 
\section{Numerical experiments \label{experiments}}
In this section, we present the results of the numerical experiment. We used six MuJoCo tasks: InvertedPendulum-v5, InvertedDoublePendulum-v5, HalfCheetah-v5, Hopper-v5, Walker2d-v5, and Ant-v5, provided by Gymnasium \cite{MuJoCo,gymnasium}. Here, we set the done signal $d=1$ only when Gymnasium determined that the environment was terminated. In other words, we did not regard that ``the environment was done'', when the environment was truncated because of the time limit. In all calculations, neither batch normalization nor observation normalization was applied. 
The performance of each simulation was evaluated every 2000 time steps. In each evaluation, we performed the simulation without adding exploration noise when $t > T_{ \mathrm{rand} }$, and chose actions randomly when $t \leq T_{ \mathrm{rand} }$. The score of each evaluation was calculated by averaging the episode rewards over 10 episodes. Furthermore, to smooth the learning curves, moving averages over 10 evaluations were computed. We took the average and standard deviation of 8 independent trials, and the latter was treated as the error. Hyperparameters used in the calculations were listed in Table~\ref{hyperparameters}.

\subsection{Confirmation that the LR update of the critic works \label{woDDPGc} }
First, we calculated the score for the case where the DDPG update of the critic was not executed, to evaluate whether the LR update of the critic was effective. In this calculation, both the DDPG and LR updates were applied to the actor. To reduce the dependency on the initial distribution of NNs, the initial DDPG update after random steps (line~\ref{DDPG_for_RDsteps} of Algorithm \ref{alg1}) was retained. Here, we calculated the case where $\beta_c$ and $\beta ' _c$ varied according to eqs.~(\ref{DLSDDPG_beta_c}) and (\ref{DLSDDPG_beta_prime_c}) and Table~\ref{hyperparameters}, and the case where these values were fixed at $\beta_c = \beta ' _c = 10^{-3}$, because the performance of the latter case in some tasks was apparently better than that of the former case.

\begin{figure*}[!tb]
 \centering
\includegraphics[width = 12.0cm]{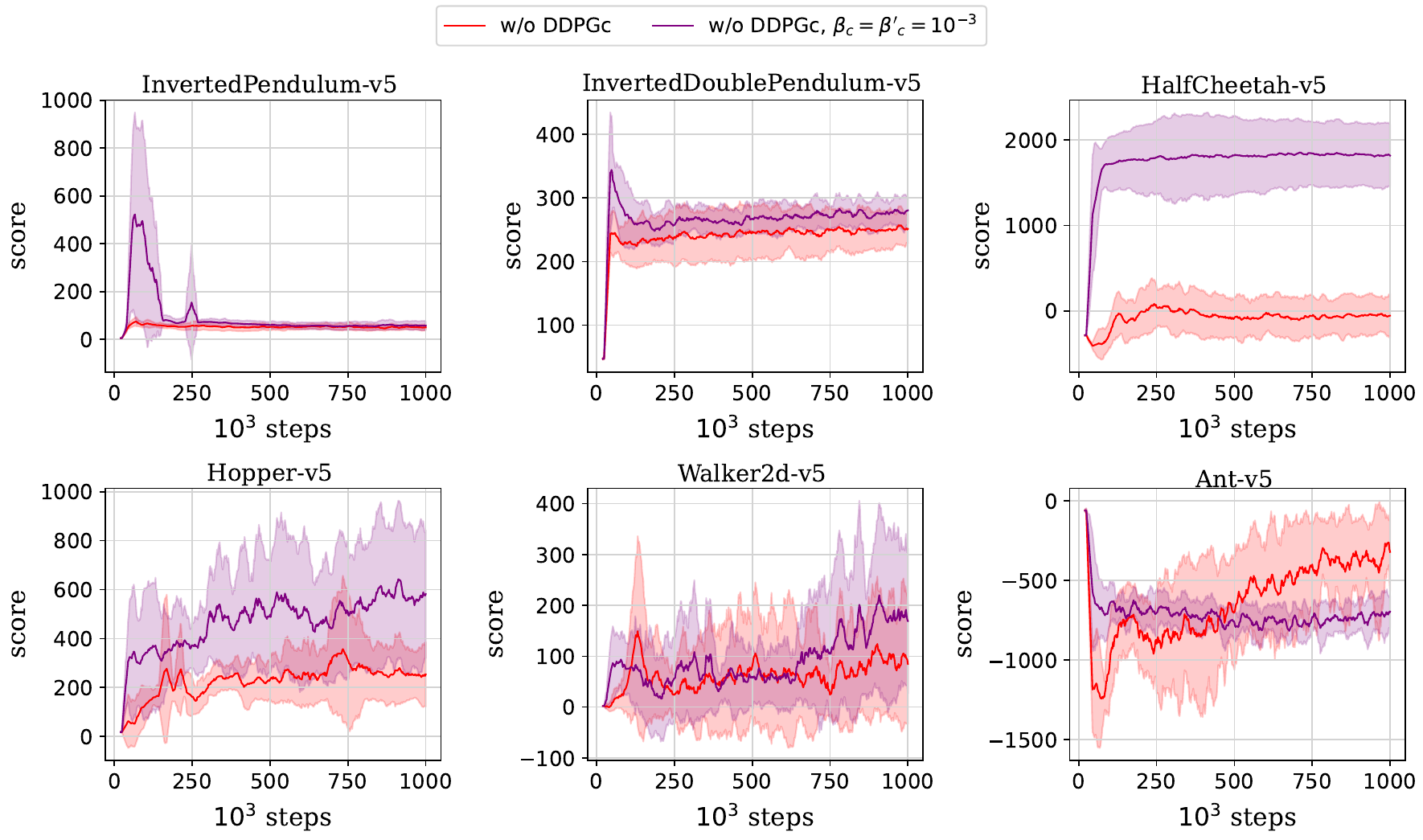}
\caption{Learning curves of the cases that did not use DDPG update for the critic, for six MuJoCo tasks. The case where $\beta_c$ and $\beta ' _c$ varied according to eqs.~(\ref{DLSDDPG_beta_c}) and (\ref{DLSDDPG_beta_prime_c}) and Table~\ref{hyperparameters} (red), and the case where $\beta_c = \beta ' _c = 10^{-3}$ (purple) were investigated. }
\label{LC_LSonlyC}
\end{figure*}

The result is shown in Fig.~\ref{LC_LSonlyC}. From this figure, it can be observed that learning progressed in some tasks, such as HalfCheetah and Hopper. However, the scores for other tasks show the unstable learning curves or no learning at all except for the initial DDPG update after random steps. Therefore, while the LR update of the critic itself is correct, it is difficult to train the NN unless it is combined with the DDPG update.
Note that we did not calculate the case where $\beta_c = \beta ' _c = 10^{-3}$ in the following sections. This is because fixing these parameters to small values made the output weight matrix $\varphi ^{out}$ vulnerable to divergence, especially when both DDPG and LR updates were used. We will discuss this issue further in Sec.~\ref{effect_regul}.

\subsection{Confirmation that the LR update of the actor works \label{woDDPGa} }
In this section, we investigated the case where the DDPG update of the actor at $t > T_{ \mathrm{rand} }$ was not executed, to examine the effect of the LR update of the actor. Note that initial DDPG update (line~\ref{DDPG_for_RDsteps} of Algorithm \ref{alg1}) was performed, as in the previous section. Regarding the actor, the effect of the OAC should also be investigated. Therefore, we calculated the scores for four cases: whether the DDPG update is applied to the actor, and whether the OAC is adopted. When the OAC is not used, we determine the action using the following equation:
\begin{equation}
a = C (\mu _{\theta} (s) + \epsilon ) . \label{AAC}
\end{equation}
Namely, the clipped output of the actor, $\mu _{\theta} (s)$, is adopted as the action instead of the optimal action, $o$, in this case. We refer to the action selection using eq.~(\ref{AAC}) as the actor action choosing (AAC). Note that in every calculation of this section, the optimal action $o$ was used as the training data for the LR update of the actor, even when AAC was adopted. Therefore, the calculation of eq.~(\ref{optim_lbfgs}) itself was required for every case. For the critic, both the DDPG and LR updates were executed in these calculations.

\begin{figure*}[!tb]
 \centering
\includegraphics[width = 12.0cm]{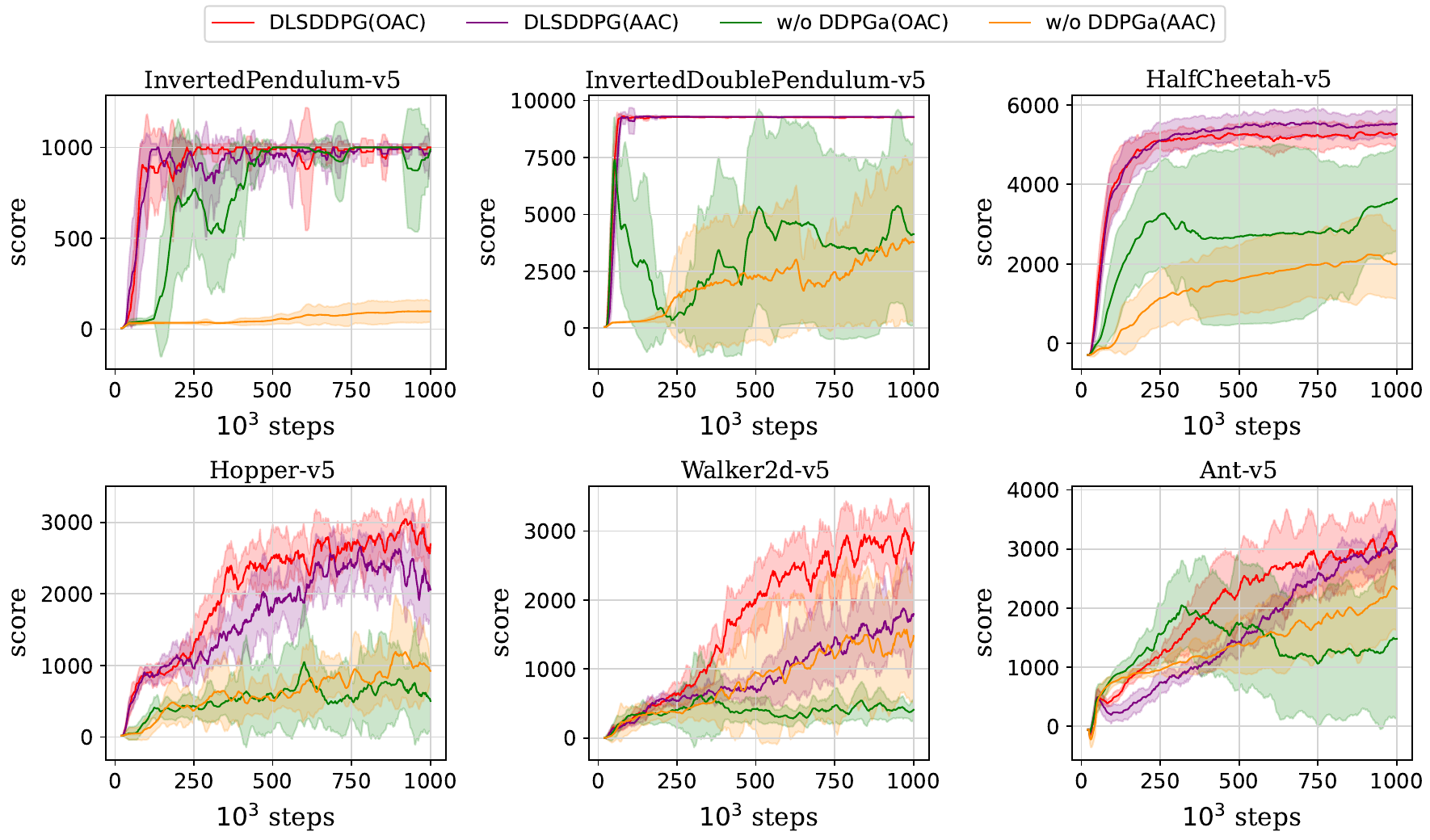}
\caption{Learning curves of DLS-DDPG with OAC(red) and AAC(purple), and the cases that did not use DDPG update for the actor with OAC(green) and AAC(orange), for six MuJoCo tasks }
\label{LC_LSonlyA}
\end{figure*}

The results are shown in Fig.~\ref{LC_LSonlyA}. Comparing the scores of OAC and AAC in this figure, we observe that OAC accelerates learning but becomes unstable without the DDPG update, especially in environments where termination is likely to occur (such as Hopper or Walker2d). Notably, the learning proceeds even in the case of AAC without the DDPG update of the actor. In this case, the only way to modify the policy is through the LR update of the actor using eq.~(\ref{DLSDDPG_actor_LR}). Therefore, we can confirm that the optimal action $o$ serves as the training data, based on this result.

\subsection{Main result \label{main_result} }

\begin{figure*}[!tb]
 \centering
\includegraphics[width = 12.0cm]{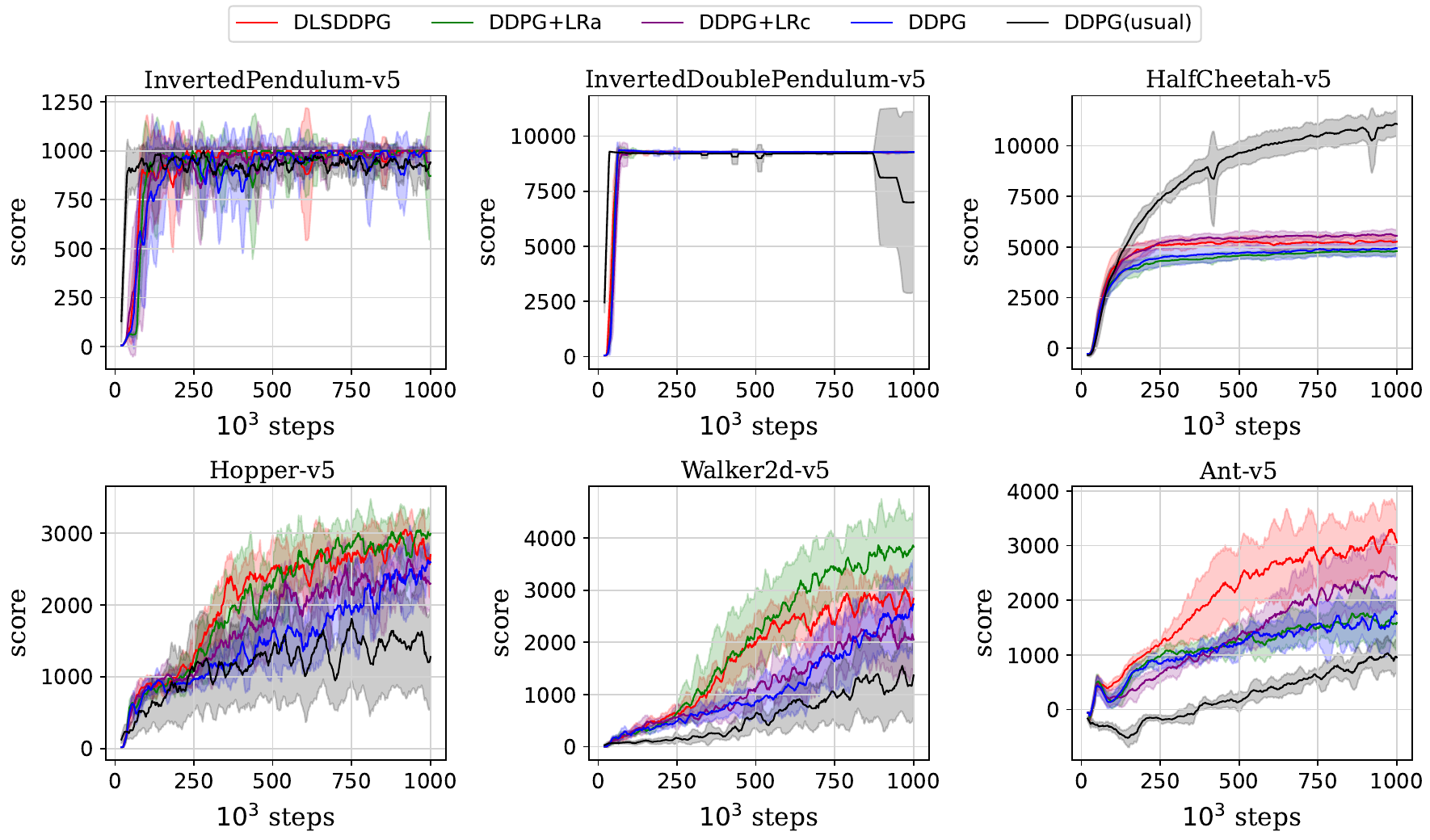}
\caption{Learning curves of DLS-DDPG(red), DDPG with LR update of actor(green) or critic(purple), our DDPG(blue) and the usual DDPG(black), for six MuJoCo tasks }
\label{LC_main}
\end{figure*}

In this section, we investigated the performance of DLS-DDPG and compared it with that of DDPG, to examine whether the LR update improves performance. When the LR update of the actor was not used, we selected the action using the AAC explained in the previous section. Additionally, the soft target updates after the LR updates using eqs.~(\ref{DLSDDPG_softc}) and (\ref{DLSDDPG_softa}) were not performed if the corresponding LR updates were not.
DDPG in our calculations differs significantly from the commonly used versions in terms of the activation functions and penalty terms, for example. Therefore, we also calculated the results for DDPG under the standard architecture, which we referred to as ``usual DDPG'' subsequently. For the hyperparameters of the usual DDPG, we used the same value as those found in the Stable Baselines3 \cite{SB3,SB3doc} and RL Zoo repository \cite{RLzoo}, to the best of our ability. Specific values for the hyperparameters of the usual DDPG are listed in the rightmost column of Table~\ref{hyperparameters}. The initial DDPG update, as described in line~\ref{DDPG_for_RDsteps} of Algorithm \ref{alg1}, was executed also in the usual DDPG.  
Fig.~\ref{LC_main} shows the learning curves for DLS-DDPG(red), DDPG with LR executed only for either the actor (green) or critic (purple), DDPG using our architectures and hyperparameters (blue), and the usual DDPG explained above (black). We also summarized the scores averaged over all evaluations in $t > 900000$ in Table~\ref{scores_table}. From these graphs and table, we can observe that in Hopper, Walker2d, and Ant, DLS-DDPG appears to learn faster than the case where the LR update is not introduced. Hence, LR update appears to promote learning in these tasks. In contrast, in HalfCheetah, the performance of our architecture is significantly worse than the usual DDPG, regardless of whether the LR update is used. We discuss this issue in the next section. Additionally, in Walker2d, the case where the LR update is introduced only to the actor shows the best performance. 
To find the exact cause of this behavior is challenging. One possible explanation is that the rapid update of the critic destabilizes the learning process, especially in difficult tasks.

\begin{table*}
\hspace{-2.0cm}
\begin{tabular}{cccccc}
 task & DLS-DDPG & DDPG+LRa & DDPG+LRc & DDPG & usual DDPG \\ \hline\hline
InvertedPendulum-v5 & 998 $\pm$ 7 & 968 $\pm$ 83 & 982 $\pm$ 27 & 973 $\pm$ 45 & 934 $\pm$ 46 \\ \hline
InvertedDoublePendulum-v5 & 9268 $\pm$ 7 & 9267 $\pm$ 4 & 9269 $\pm$ 18 & 9268 $\pm$ 11 & 7518 $\pm$ 3342 \\ \hline
HalfCheetah-v5 & 5269 $\pm$ 276 & 4780 $\pm$ 200 & 5596 $\pm$ 223 & 4903 $\pm$ 379 & \textbf{10831 $\pm$ 608} \\ \hline
Hopper-v5 & \textbf{2849 $\pm$ 171} & \textbf{2896 $\pm$ 193} & 2386 $\pm$ 246 & 2444 $\pm$ 252 & 1431 $\pm$ 594 \\ \hline
Walker2d-v5 & 2833 $\pm$ 298 & \textbf{3760 $\pm$ 600} & 2111 $\pm$ 655 & 2532 $\pm$ 688 & 1251 $\pm$ 679 \\ \hline
Ant-v5 & \textbf{3084 $\pm$ 586} & 1602 $\pm$ 400 & \textbf{2453 $\pm$ 591} & 1616 $\pm$ 427 & 960 $\pm$ 263 \\ \hline
\end{tabular}
\vspace{1.0mm}
\caption{Evaluated scores averaged over $t > 900000$. Here, DDPG with LR executed only for actor (critic) is abbreviated as DDPG+LRa (DDPG+LRc). For HalfCheetah, Hopper, Walker2d, and Ant, the best scores and those that do not differ from them within the range of error are written in bold. }
\label{scores_table}
\end{table*}

\begin{figure*}[!tb]
 \centering
\includegraphics[width = 12.0cm]{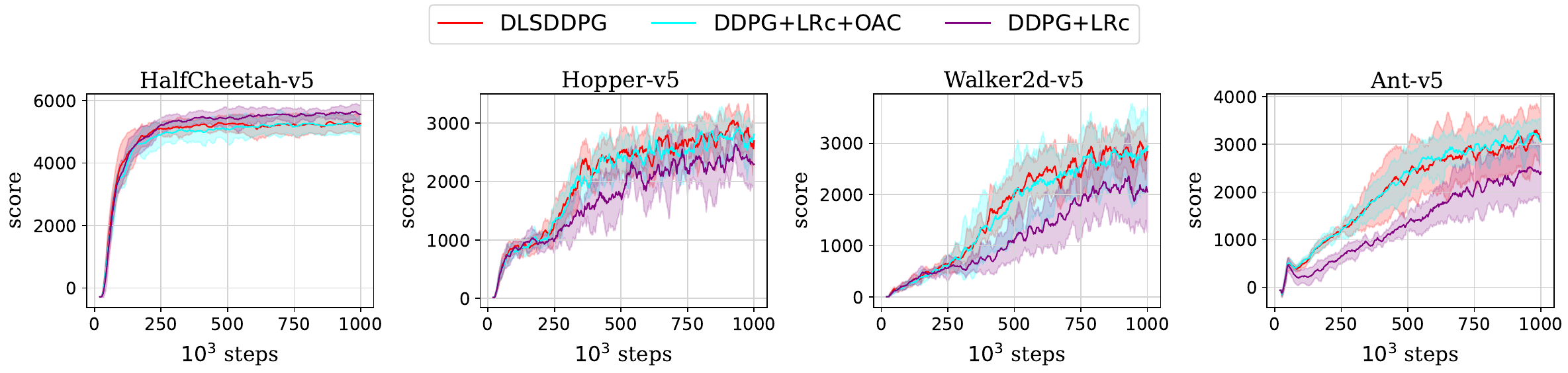}
\caption{Learning curves of DLS-DDPG(red), DDPG with LR update of the critic and OAC (cyan), and that with LR update of the critic and AAC (purple), for four MuJoCo tasks }
\label{LC_OACwoLSA_LSC}
\end{figure*}

\begin{figure*}[!tb]
 \centering
\includegraphics[width = 12.0cm]{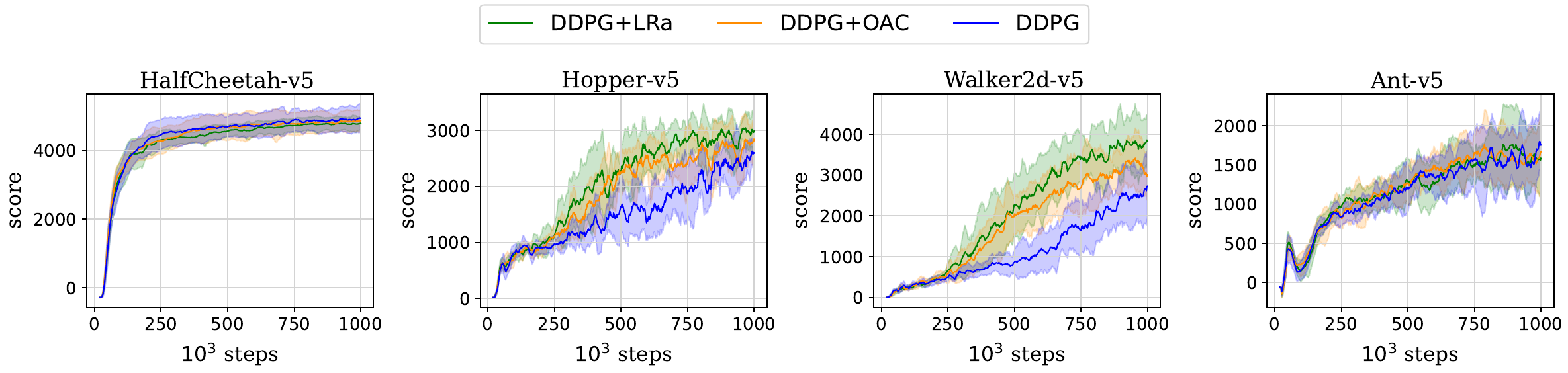}
\caption{Learning curves of DDPG with LR update of the actor and OAC (green), DDPG with OAC (orange), and that with AAC (blue), for four MuJoCo tasks }
\label{LC_OACwoLSA_noLSC}
\end{figure*}

As previously mentioned, Fig.~\ref{LC_main} and Table~\ref{scores_table} do not show the cases where OAC was adopted but the LR update of the actor was not applied. Therefore, we also calculated the performance of such cases to evaluate whether the LR update of the actor or the OAC had significant contribution to the performance improvement. Fig.~\ref{LC_OACwoLSA_LSC} compares the results of three methods: DLS-DDPG (red), DDPG with LR update of the critic and OAC (cyan), and that with LR update of the critic and AAC (purple). Similarly, Fig.~\ref{LC_OACwoLSA_noLSC} compares the calculations without the LR update of the critic. Specifically, it shows three cases: DDPG with LR update of the actor and OAC (green), DDPG with OAC (orange), and that with AAC (blue). Note that the red, purple, green and blue curves of these figures represent the same meaning  as those in Fig.~\ref{LC_main}. In Figs.~\ref{LC_OACwoLSA_LSC} and \ref{LC_OACwoLSA_noLSC}, the results for InvertedPendulum and InvertedDoublePendulum are abbreviated, as the techniques proposed in this study have little impact on their performance. According to Figs.~\ref{LC_OACwoLSA_LSC} and \ref{LC_OACwoLSA_noLSC}, in most cases where the performance is improved by the proposed method, such improvement is mainly due to OAC, while the LR update of the actor has a relatively smaller effect. In particular, the difference between DLS-DDPG and DDPG with the LR update of the critic and OAC is negligible, as shown in Fig.~\ref{LC_OACwoLSA_LSC}. In this case, quasi-Newton method is considered capable of calculating the argmax of $Q(s,a)$ with high accuracy and the effect of the LR update of the actor is relatively slight.

\subsection{Effect of the regularization terms \label{effect_regul} }
As mentioned earlier, our architecture differs from the usual DDPG in several aspects. Among these differences, the activation function or the number of the hidden layers can be adjusted from our architecture, provided that the form of $\nabla _a Q$ is accordingly modified from that in eq.~(\ref{fprime_Q}). Conversely, the regularization terms are essential for DLS-DDPG to prevent the divergence of the weight matrices. 
In this section, we examine the effect of these regularization terms. While they sometimes hinder the adaptation to the tasks, they can also prevent the overfitting.

We first calculated $\log _{10} n_{\theta}$ and $\log _{10} n_{\varphi}$, the logarithms of  the normalized Frobenius norms of the output weight matrices, while varying the value of the coefficients $\beta_a$, $\beta ' _a$, $\beta_c$, and $\beta ' _c$. This was done to investigate whether weakening the regularization terms result in the divergence of these matrices. Here, we refer to the case where these coefficients change according to eqs.~(\ref{DLSDDPG_beta_a})--(\ref{DLSDDPG_beta_prime_c}) and Table~\ref{hyperparameters} as the default case, and compare it with the cases where they are fixed to constant values. Additionally, calculations were executed also for the case where the DDPG update of the corresponding network at $t > T_{ \mathrm{rand} }$ did not exist, to investigate whether combining the DDPG and LR updates increase these norms. The actual investigation was executed for the HalfCheetah-v5 and Hopper-v5. Note that the investigation on $\beta_a$ and $\beta ' _a$ was executed only in the calculation of $\log _{10} n_{\theta}$, and that on $\beta_c$ and $\beta ' _c$ was executed only in the calculation of $\log _{10} n_{\varphi}$. Namely, only coefficients that directly affect the output weight matrix under consideration were investigated. For other coefficients, default changes were executed. 

\begin{figure*}[!tb]
 \centering
\includegraphics[width = 11.0cm]{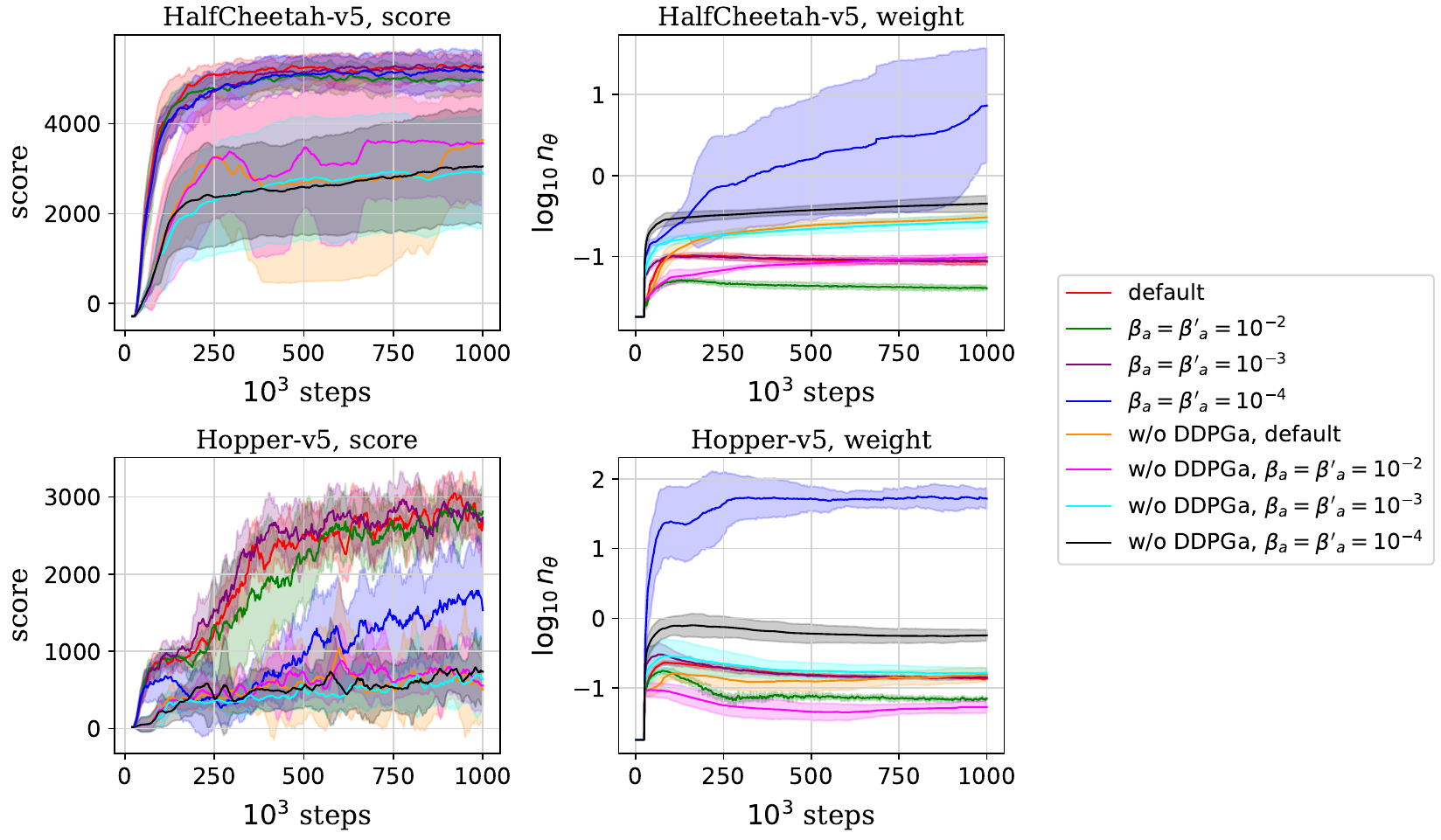}
\caption{Effect of the regularization terms of the actor on DLS-DDPG, calculated for the HalfCheetah-v5 and Hopper-v5. The left and right columns mean the score and logarithm of the magnitude of the output weight, $\log _{10} n_{\theta}$, respectively. The investigated cases are as follows:
DLS-DDPG with default case (red), $\beta_a = \beta ' _a = 10^{-2}$ (green), $10^{-3}$ (purple), and $10^{-4}$ (blue), DLS-DDPG without the DDPG update of the actor with default case (orange), $\beta_a = \beta ' _a = 10^{-2}$ (magenta), $10^{-3}$ (cyan), and $10^{-4}$ (black). }
\label{small_regulA}
\end{figure*}

\begin{figure*}[!tb]
 \centering
\includegraphics[width = 11.0cm]{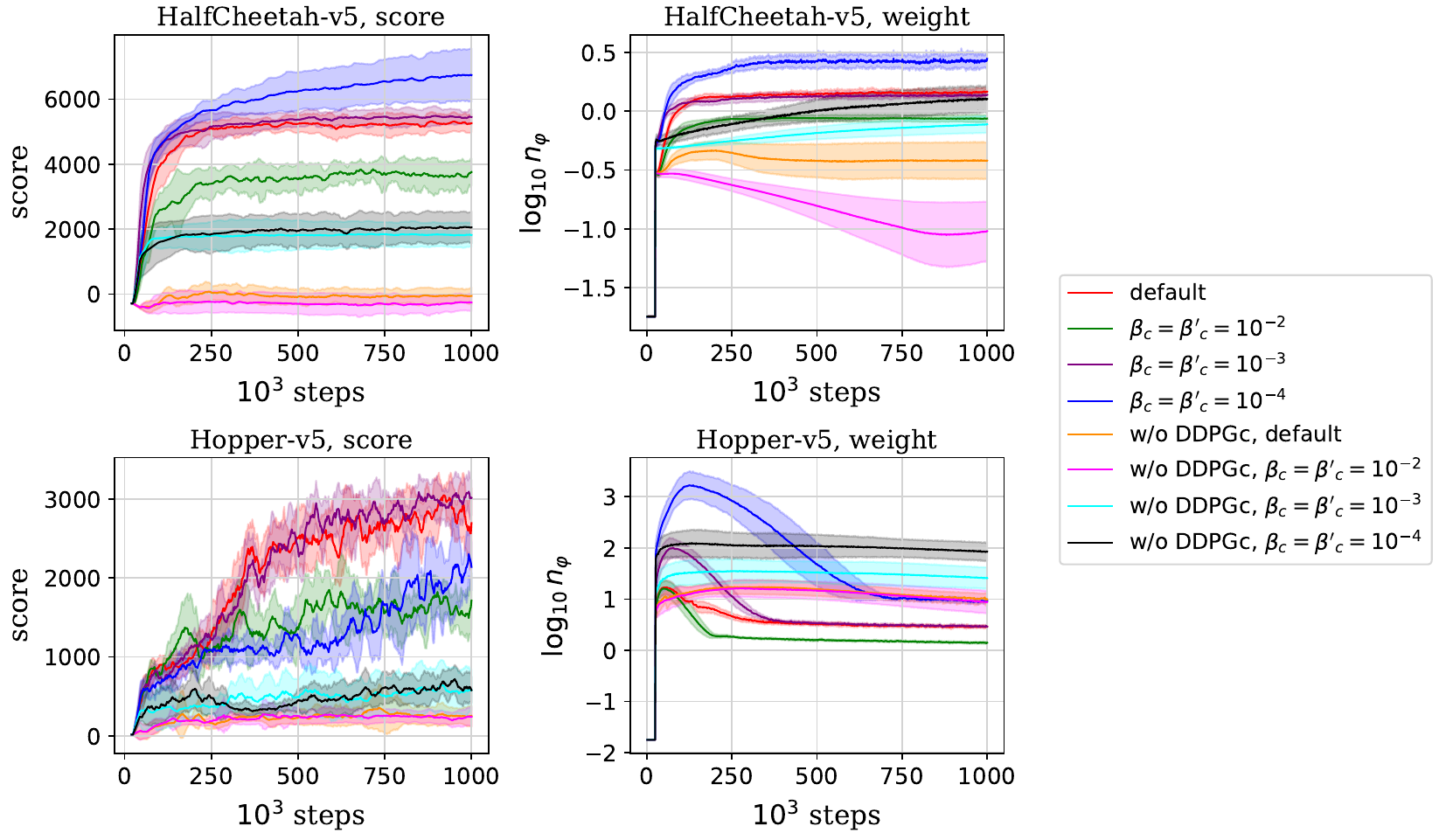}
\caption{Effect of the regularization terms of the critic on DLS-DDPG, calculated for the HalfCheetah-v5 and Hopper-v5. The left and right columns mean the score and logarithm of the magnitude of the output weight, $\log _{10} n_{\varphi}$, respectively. The investigated cases are as follows:
DLS-DDPG with default case (red), $\beta_c = \beta ' _c = 10^{-2}$ (green), $10^{-3}$ (purple), and $10^{-4}$ (blue), DLS-DDPG without the DDPG update of the critic with default case (orange), $\beta_c = \beta ' _c = 10^{-2}$ (magenta), $10^{-3}$ (cyan), and $10^{-4}$ (black). }
\label{small_regulC}
\end{figure*}

The results are shown in Figs.~\ref{small_regulA} and \ref{small_regulC}. Note that we did not apply the moving average to $n_{\theta}$ and $n_{\varphi}$. From these figures, it can be observed that $n_{\theta}$ and $n_{\varphi}$ tend to be larger when the DDPG update is executed. This indicates that combining the two different updates makes the output weight matrices vulnerable to the divergence, highlighting the importance of controlling the regularization terms as in eqs.~(\ref{DLSDDPG_beta_a})--(\ref{DLSDDPG_beta_prime_c}). Indeed, in the case of Hopper, for example, the output weight matrix of the critic in the default case follows a more moderate curve than the case when $\beta_c = \beta ' _c = 10^{-3}$ (See the bottom right graph of Fig.~\ref{small_regulC}).

To assess the performance in the complete absence of regularization terms, we next calculated the score of our architecture when the LR updates (and corresponding soft target updates) were not introduced, dropping the regularization terms. Specifically, new calculations were conducted for the following three cases: $\beta_a = \beta ' _a = 0$, $\beta_c = \beta ' _c = 0$, and $\beta_a = \beta ' _a = \beta_c = \beta ' _c = 0$. Here, the four parameters $\beta_a$, $\beta ' _a$, $\beta_c$, and $\beta ' _c$, followed the eqs.~(\ref{DLSDDPG_beta_a})--(\ref{DLSDDPG_beta_prime_c}) with the hyperparameters listed in Table~\ref{hyperparameters}, as in the previous sections, unless they were set to 0.

\begin{figure*}[!tb]
 \centering
\includegraphics[width = 12.0cm]{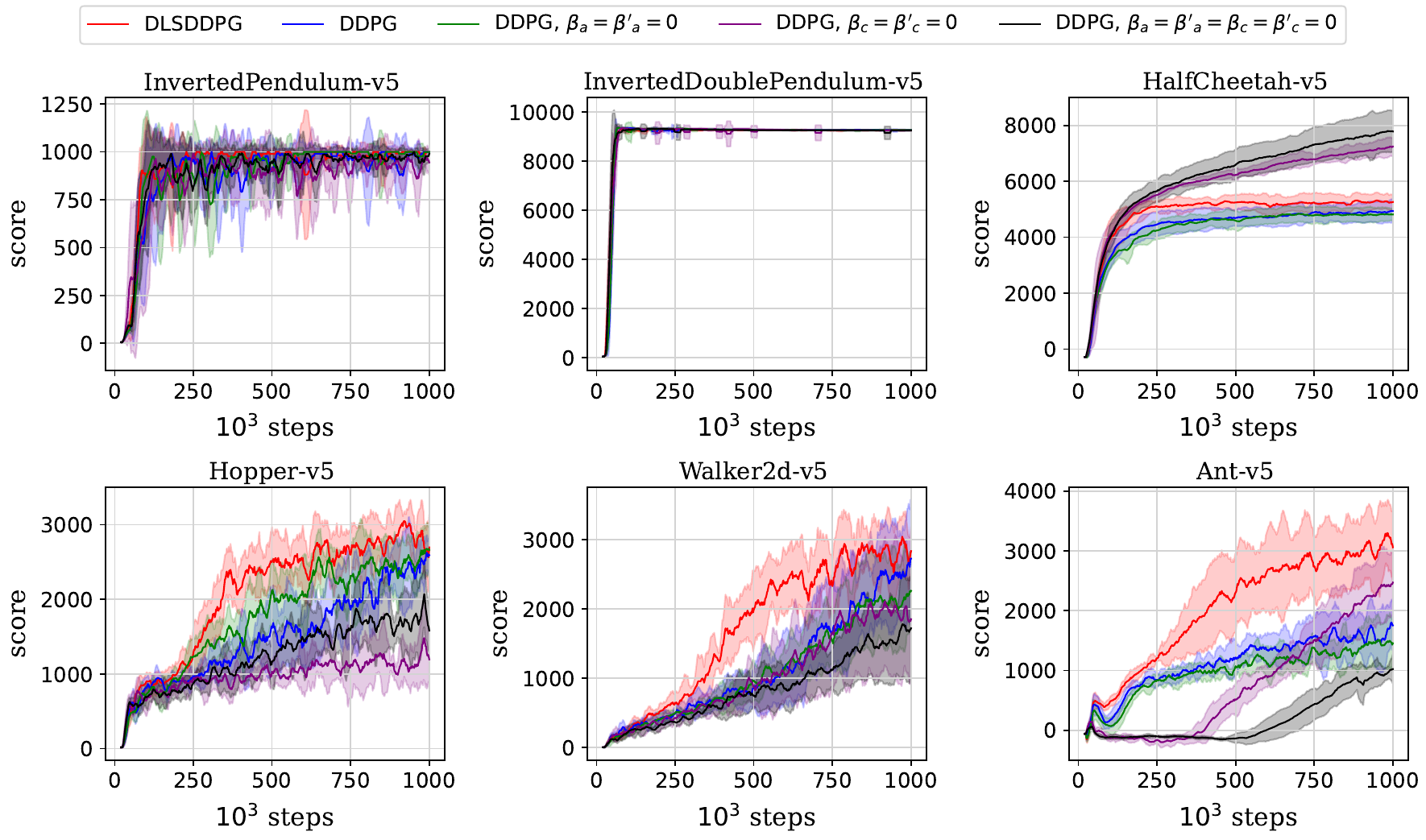}
\caption{Learning curves of DDPG with $\beta_a = \beta ' _a = 0$ (green), $\beta_c = \beta ' _c = 0$ (purple), and $\beta_a = \beta ' _a = \beta_c = \beta ' _c = 0$ (black), for six MuJoCo tasks. The red and blue curves are the same as that of Fig.~\ref{LC_main}, the scores of DLS-DDPG and DDPG when all regularization terms exist. }
\label{LC_no_regul}
\end{figure*}
The result is shown in Fig.~\ref{LC_no_regul}. From this figure, we can observe that the performances of the cases without regularization terms show a similar tendency to the usual DDPG in the previous section. Specifically, the score of HalfCheetah under $\beta_c = \beta ' _c = 0$ continues to increase, although its value remains lower than that of the usual DDPG. In contrast, the scores for Hopper, Walker2d, and Ant tend to worsen when the regularization terms are removed. Therefore, the regularization terms are believed to contribute significantly to the performance differences between the usual DDPG and DDPG under our architecture.

\section{Summary \label{summary} }

In this study, we proposed the DLS-DDPG method, which combines DDPG method with LR update and the quasi-Newton method. The LR update of the critic was similar to that of FQI, except for the use of the target network. In contrast, for the actor, the loss function of the DDPG cannot be applied to the LR update. Instead, we calculated the optimal action $o$ by applying the quasi-Newton method to the action-value function $Q(s,a)$. This value $o$ was used both as the agent's action and as the training data for the LR update of the actor. Numerical experiments showed that the proposed method improved the performance of DDPG in some MuJoCo tasks. Here, when the DDPG update of the actor exists, the effect of using the optimal action is greater than that of the LR update of the actor.

As explained in Sec.~\ref{review_DDPG}, there are several improved variants of DDPG, such as TD3 and SAC, which are widely used today \cite{Fujimoto18, Haarnoja18_1, Haarnoja18_2}. Therefore, our next objective is to apply our method to these improved variants. While TD3 and SAC differ in some details, they share a key similarity in that both adopt the clipped double-Q trick. This technique is inspired by double Q-learning \cite{Hasselt10} and uses two critic networks to prevent the overestimation of $Q(s,a)$. Moreover, more recent algorithms, such as the Truncated Quantile Critics (TQC) \cite{Kuznetsov20} and Randomized Ensembled Double Q-learning (REDQ) \cite{Chen21} use more critic networks. Hence, to combine our method with these variants, we should first investigate which of the two or more critic networks should be used to calculate the optimal action. The solution to this problem is not obvious and will require careful investigation.

For the practical application of our method, the vulnerability to the divergence of the output weight matrices remains a concern. In this study, we controlled the coefficients for the regularization terms, $\beta_a$, $\beta ' _a$, $\beta_c$, and $\beta ' _c$, to mitigate this problem. However, introducing these terms resulted in worse performance in some tasks, such as HalfCheetah. Moreover, the existence of these coefficients made hyperparameter tuning more challenging. Developing methods to address these challenges will be a focus of the future work. Conversely, in previous studies combining LR and DRL methods, regularization terms themselves were adopted, but the divergence of weight matrices was not reported \cite{Levine17,Chung19,Li24}. Therefore, it should be investigated whether cases involving continuous action or a DDPG-like architecture requires particularly careful tuning of regularization terms.

The source code for this work is available at \url{https://github.com/Hisato-Komatsu/DLS-DDPG/}.

\section*{Declaration of Competing Interest }
The author declares that there are no known competing financial interests or personal relationships that could have influenced the work reported in this paper. 

\section*{Acknowledgements}
This study is supported by the Grant for Startup Research Project of Shiga University.
We would like to thank Editage for their assistance with English language editing.

\section*{Declaration of Generative AI and AI-assisted technologies in the writing process}
During the preparation of this work the author used DeepL and Microsoft Copilot in order to improve the English language. After using these tools, the author reviewed and edited the content as needed and takes full responsibility for the content of the publication.

\end{document}